\documentclass[letterpaper, 10 pt, conference]{ieeeconf}
\IEEEoverridecommandlockouts
\makeatletter
\let\NAT@parse\undefined    % These 3 lines are for the natbib package
\makeatother

\usepackage[numbers]{natbib}    % Alternative to the cite package
\usepackage[pagebackref=true]{hyperref}    % To make in-text citations link to references: https://tex.stackexchange.com/questions/247104/hyperref-doesnt-link-cite-command
\usepackage{amsmath,amssymb,amsfonts}
\usepackage{algorithmic}
\usepackage{graphicx}
\usepackage{textcomp}
\usepackage{xcolor}
\usepackage[table]{xcolor}
\usepackage{tikz}
\usepackage{varwidth}
\usepackage{stfloats}
\usepackage{subcaption}   
\usepackage{scalefnt}
\usepackage{graphicx}
\usepackage{makecell}    % Gives \makecell command for new lines in table cells
\usepackage{multirow}
\usepackage{booktabs, multirow} 
\usepackage{pifont}
\usepackage{soul}
\usepackage{changepage}
\usepackage{floatrow}
\usepackage[T1]{fontenc} 
\usepackage{caption}        % defines \captionlabel (needs v3.x or later) :contentReference[oaicite:0]{index=0}

\usepackage{xcolor} % already included in your file, but keep if unsure

\usetikzlibrary{backgrounds,calc,fit,positioning,shapes.symbols}
\def\BibTeX{{\rm B\kern-.05em{\sc i\kern-.025em b}\kern-.08em
    T\kern-.1667em\lower.7ex\hbox{E}\kern-.125emX}}
\begin{document}

\title{\LARGE \bf
Flexible and Efficient Spatio-Temporal Transformer for Sequential Visual Place Recognition
}

% Put authors here after acceptance
\author{Yu Kiu (Idan) Lau, Chao Chen, Ge Jin, Chen Feng\textsuperscript{\ding{41}}\\
% New York University\\
% \texttt{\small \{idanlau, cc7287, gj2148, cfeng\}@nyu.edu}
% \thanks{The work was supported in part through NSF grants 2238968, 2322242, 2121391, and 2036870. We thank Xuchu Xu for insightful discussions.}
% \thanks{* Equal contribution}
% \thanks{\ding{41} Corresponding author.}% <-this % stops a space
\thanks{\ding{41}Corresponding author. All authors are with the Center for Robotics and Embodied Intelligence (CREO) at New York University, Brooklyn, NY 11201, USA. \texttt{\small cfeng@nyu.edu} }
\thanks{The work was supported in part through NSF grants 2238968, 2322242, and 2514030.}
% Brooklyn, NY 11201, USA  {\tt\small cfeng@nyu.edu}}
% \thanks{$^{2}$Li Ding is with University of Rochester, Rochester, NY 14627, USA {\tt\small l.ding@rochester.edu}}
% \thanks{The webpage of this paper is available at https://ai4ce.github.io/TF-VPR/.}
% \thanks{Digital Object Identifier(DOI): see top of this page.}
}
% \thanks{The work was supported in part through NSF grants 2238968, 2322242, 2121391, and 2036870.}
\maketitle 

\begin{abstract}
Sequential Visual Place Recognition (Seq-VPR) leverages transformers to capture spatio-temporal features effectively. In practice, a transformer-based Seq-VPR model should be flexible to the number of frames per sequence (seq-length), deliver fast inference, and have low memory usage to meet real-time constraints. However, existing approaches prioritize performance at the expense of flexibility and efficiency. To address this gap, we propose \textbf{Adapt-STformer}, a Seq-VPR method built around our novel \emph{Recurrent Deformable Transformer Encoder (Recurrent-DTE)}, which uses an iterative recurrent mechanism to fuse information from multiple sequential frames. This design naturally supports variable seq-lengths, fast inference, and low memory usage. Experiments on the Nordland, Oxford, and NuScenes datasets show that Adapt-STformer boosts recall by up to 17\% while reducing sequence extraction time by 36\% and lowering memory usage by 35\% relative to our best comparable baseline. Our code is released at https://ai4ce.github.io/Adapt-STFormer/.
\end{abstract}

\section{Introduction}
Sequential Visual Place Recognition (Seq-VPR) extends traditional VPR by leveraging multiple consecutive frames to construct more robust descriptors for a given location. By incorporating temporal information, Seq-VPR produces more consistent outputs, which is especially beneficial under challenging conditions such as low light or illumination changes \cite{Garg_2021_seqnet, Li_2022_bevformer}.  
Methods for exchanging and aggregating spatial-temporal features across frames generally fall into two categories. The first type neglects complex cross-frame temporal interactions. For example, JIST collapses per-frame embeddings into a single descriptor via learnable mean pooling, without direct frame-to-frame interaction \cite{Garg_2021_seqnet, Mereu_2022_seqvlad, Berton_2024_jist}. The second type draws inspiration from video understanding \cite{Bertasius_2021_timesformer, Feichtenhofer_2017_multiplier, cherian2020vidcap}, where recent Seq-VPR methods integrate transformer modules to model both inter-frame (temporal) and intra-frame (spatial) interactions.  

\begin{figure}[t]
    \centering
    \includegraphics[width=1\columnwidth]{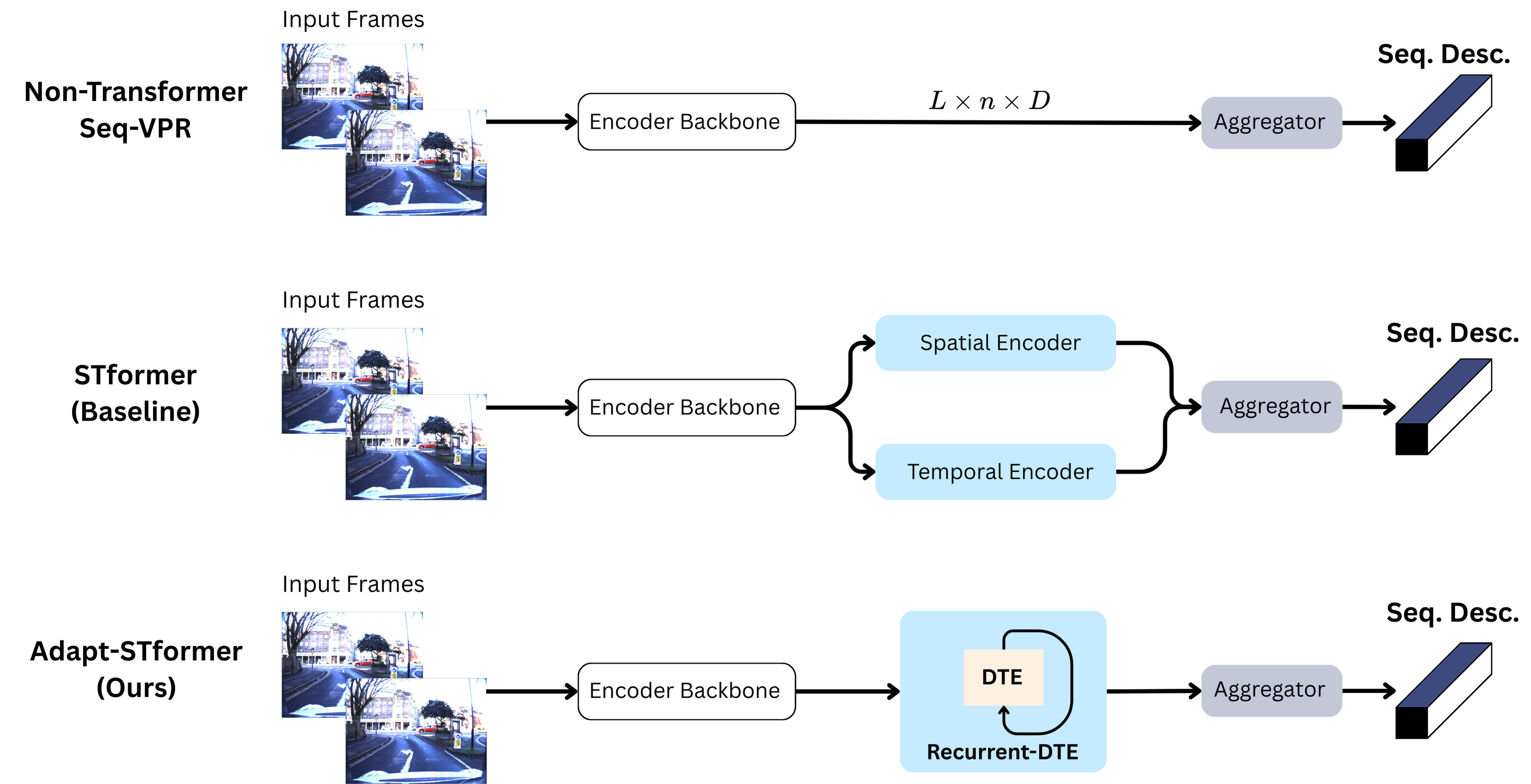}
    \caption{Comparison of Seq-VPR architectures. Top: Non-transformer methods feed backbone features directly to an aggregator. Middle: STformer handles spatial and temporal modeling with separate encoders before aggregation. Bottom: The proposed Adapt-STformer unifies spatio-temporal modeling in the Recurrent-DTE module before aggregation.}
    \label{fig:teaser}
\end{figure}

However, existing transformer‐based Seq-VPR research prioritizes performance over flexibility and efficiency. In practice, a Seq-VPR model should be flexible with respect to the number of frames per sequence (seq-length). It should also be efficient in inference speed and memory usage to meet the real-time constraints of VPR systems. At present, we believe no existing transformer-based Seq-VPR model is both efficient and flexible. Some approaches use a standard transformer layer to learn temporal interactions after extracting spatial features. However, as each additional frame introduces more tokens, the transformer’s computation grows quadratically. STformer \cite{Zhao_2024_attention} reduces computation by limiting temporal attention to local windows across frames. While this windowing lowers costs, it is not flexible with respect to seq-length. This constraint reduces practical usability: when frames are missing/distorted, the model can only rely on padding/dropping to satisfy the required seq-length. Moreover, STformer keeps separate spatial and temporal modules, adding significant overhead, leaving STformer neither flexible nor efficient.

To address this gap, we propose Adapt-STformer, a Seq-VPR framework centered on our novel Recurrent Deformable Transformer Encoder (Recurrent-DTE). Unlike prior works that process multiple frames in parallel~\cite{Schmidt_2019_RNNintro, Facil_2019_condition}, our approach unifies spatio–temporal fusion within a single holistic module. The Recurrent-DTE recurrently applies a lightweight Deformable Transformer Encoder (DTE) \cite{Zhu_2021_deformable} to frames in temporal order, from the earliest to the most recent. This design naturally encodes temporal dependencies, supports arbitrary seq-lengths, and achieves a favorable balance of flexibility, efficiency, and performance. We highlight the main contributions of this work as follows:

\begin{itemize}
    \item We present Adapt-STformer, a Seq-VPR architecture that eliminates the restriction of fixed seq-length through a lightweight unified framework, thereby achieving both flexibility and efficiency.
    
    \item We introduce Recurrent-DTE, a recurrent mechanism embedded in a DTE that iteratively processes frames in temporal order, enabling spatio-temporal modeling within a unified module. This is an effective recombination of well-known components: recurrent sequence modeling from time-series learning and DTE from object detection, for seq-VPR.
    
    \item We conduct extensive experiments on the Nordland, Oxford, and NuScenes datasets, where Adapt-STformer achieves balanced improvements, increasing recall by 10\% on average while reducing inference latency by 36\% and lowering memory usage by 35\% relative to our best comparable baseline.
\end{itemize}

\section{Related Works}

\subsection{Classic Seq-VPR}
Early Seq-VPR methods relied on handcrafted descriptors combined across frames, but these lacked robustness to environmental changes \cite{Milford_2012_seqslam}. The introduction of learning-based approaches, particularly CNN-based models, marked a shift toward data-driven sequence aggregation. SeqNet \cite{Garg_2021_seqnet} applied a lightweight 1D temporal convolution to merge a sequence into a single compact descriptor. Building on this idea, SeqVLAD \cite{Mereu_2022_seqvlad} extended NetVLAD \cite{Arandjelovic_2016_netvlad} by pooling features from all frames into a unified descriptor set and applying VLAD’s soft-assignment and residual aggregation, producing a fixed-length embedding adaptable to different sequence lengths. With a focus on efficiency, JIST \cite{Berton_2024_jist} proposed SeqGeM, a learnable mean pooling mechanism for compact sequence representation. While effective, these non-transformer Seq-VPR methods model temporal information only through simple aggregation, limiting their ability to capture complex spatio-temporal dependencies and leading to performance drops in challenging conditions \cite{Zhao_2024_attention}. Our work seeks to retain their efficiency and flexibility while addressing these limitations.

\subsection{Transformer-based Seq-VPR}
With the rise of transformers in vision tasks ~\cite{Feichtenhofer_2017_multiplier, cherian2020vidcap}, Seq-VPR methods have also adopted them to better model spatio-temporal dependencies. STFormer \cite{Zhao_2024_attention} introduced a two-stage design with a spatial transformer for frame-level features and a sliding-window temporal transformer for sequence modeling, achieving strong performance but limited by fixed sequence lengths and high computational cost due to separate encoders and dense attention. More recently, CaseVPR \cite{10884025} leveraged a DINOv2 backbone \cite{Oquab_2023_dinov2} and correlation-aware embedding, reaching SOTA results with seq-length flexibility, but at the expense of heavy computation, slow inference, and large memory use. In short, existing transformer-based Seq-VPR methods trade flexibility and/or efficiency for performance.

\subsection{Computationally Efficient Transformer}
Deformable Transformers, proposed in \cite{Zhu_2021_deformable}, employ deformable attention, where each query attends to a sparse set of learned offsets, thereby reducing the quadratic cost of global self-attention on high-resolution features. Xu et al. \cite{Xu_2023_bev360} demonstrated the effectiveness of Deformable Transformer Encoders (DTE) for multi-camera VPR, although without modeling temporal features. BEVFormer \cite{Li_2022_bevformer} integrates temporal cues by recurrently fusing past BEV features and further employs a Temporal Self-Attention (TSA) layer in addition to its DTE for spatial lookup across multiple camera views. In contrast, our design unifies spatial and temporal reasoning directly within a DTE, avoiding the need for a separate TSA layer. We found that replicating BEVFormer's additional TSA increases inference time and memory usage while not improving performance. Beyond these architectural differences, BEVFormer assumes a synchronized multi-camera setup with known camera intrinsics. In contrast, we target a practical monocular front-view scenario without access to calibration parameters, making our method applicable to real-world settings.

\subsection{Computation effcient backbone}
The choice of encoder backbone is crucial for our method's performance and efficiency. Early Seq-VPR methods commonly use encoder backbones such as ResNet \cite{he2016resnet} valued for their computational efficiency but limited in representational power. In contrast, modern Vision Transformer (ViT) backbones \cite{Oquab_2023_dinov2, lu2024cricavpr} offer rich, high-quality visual features learned from massive self-supervised training, yet they incur substantial computational cost. Compact Convolutional Transformer (CCT) \cite{Hassani_2021_compact} represents a hybrid of convolutional tokenization and transformer layers, showing promising performance while remaining relatively lightweight compared to pure ViT \cite{Dosovitskiy_2021_vit} counterparts. Motivated by this balance of efficiency and performance, we adopt CCT as the encoder backbone for our Seq-VPR framework.

\begin{figure*}[t]
    \centering
    \includegraphics[width=\textwidth, height=10cm, keepaspectratio=true]{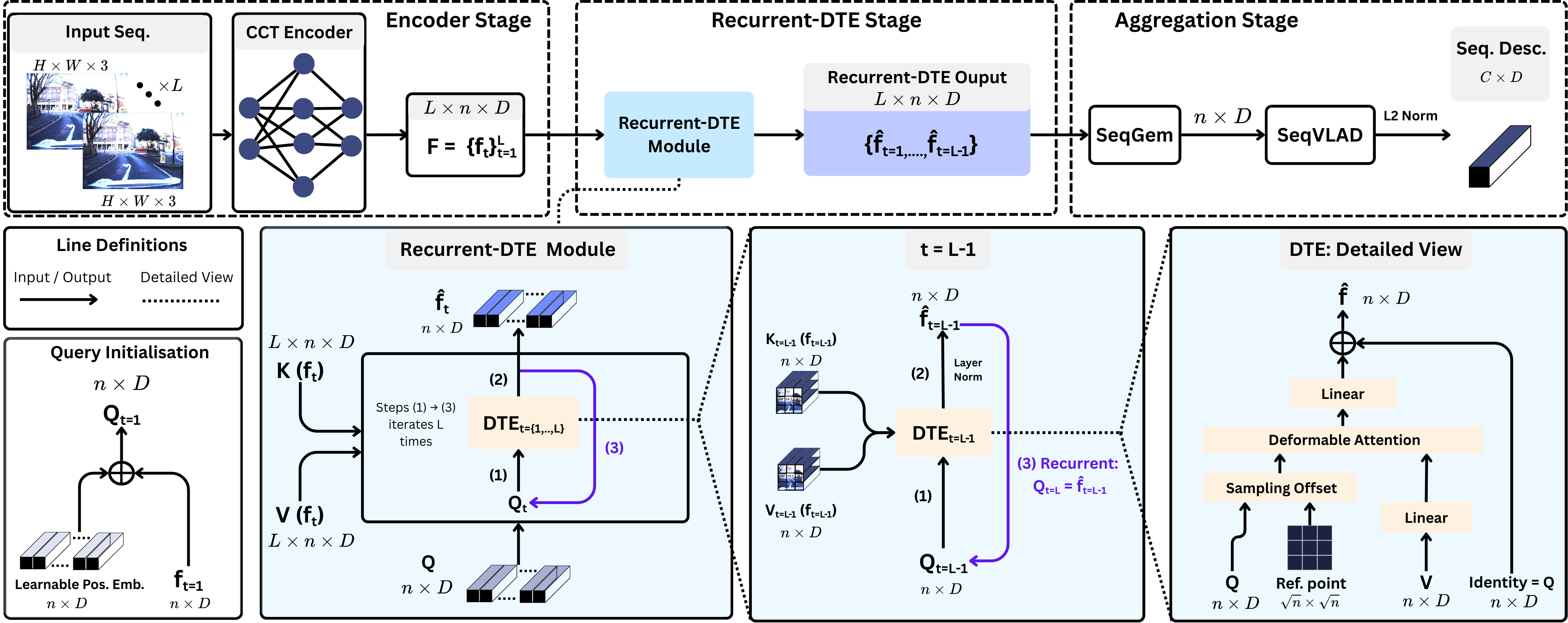}
    \caption{\textbf{Proposed architecture of Adapt-STformer.} 
    A frame sequence $S$ enters the \textbf{Encoder Stage}, 
    where the CCT384 backbone tokenizes each input frame, producing 
    $F=\{f_t\}_{t=1}^{L}$. 
    \textbf{Recurrent-DTE Stage} performs \emph{iterative recurrent} spatio-temporal modeling on $F$: 
    at $t$, the previous DTE output provides  $Q_t=\hat{f}_{t-1}$ and the current frame supplies $K_t=V_t=f_t$, 
    with $Q_{t=1}=f_{t=1}+\Delta$; this recurrence yields $\{\hat{f}_t\}_{t=1}^{L}$. 
    \textbf{Aggregation Stage} stacks $\{\hat{f}_t\}$ into a tensor $\hat{F}$, applies SeqGeM to obtain 
    $\tilde F$, and processes it with SeqVLAD into the final sequential descriptor.}

    \label{fig:adapt_architecture}
\end{figure*}

\section*{Methodology}

\begin{table}[h]
\centering
\captionsetup{font=sc}
\captionsetup{font={scriptsize, sc, stretch=1.3}, justification=centering}
\caption{Summary of major notations for Adapt-STformer.}
\label{tab:notation}
\footnotesize 
\renewcommand{\arraystretch}{0.95}
\setlength{\tabcolsep}{5pt}
\begin{tabular}{>{\columncolor{gray!10}}l l}
\toprule
$L$ & Number of frames/seq-length \\
$t$ & Iteration/frame index, $t \in \{1,\dots,L\}$ \\
$f_t$ & The encoder output feature from the $t$th image \\
$n$ & Number of tokens in image feature $f_t$ \\
$C$ & Number of VLAD clusters \\
$D$ & Token embedding dimension \\
$\hat{f}_t$ & DTE output at $t$ \\
$F$ & The set of encoder outputs across the sequence \\
$(Q_t, K_t, V_t)$ & Query, Key, and Value at iteration $t$ \\
\bottomrule
\end{tabular}
\end{table}

\subsection*{A. Overall Architecture }
Adapt-STformer takes as input a frame sequence $S=\{s_i\}_{i=1}^L$, with each $s_i\in\mathbb{R}^{H\times W\times 3}$, and outputs a NetVLAD descriptor $V\in\mathbb{R}^{C\times D}$. There are three intermediate stages between input and output.

\begin{itemize}

\item \textbf{Encoder Stage:} Encode $S$ with CCT to obtain $F=\{f_t\}_{t=1}^L \in \mathbb{R}^{L\times n\times D}$.  
\item \textbf{Recurrent-DTE Stage:} Refine $F$ iteratively via Recurrent-DTE, producing $\hat{F}=\{\hat{f}_t\}_{t=1}^L \in \mathbb{R}^{L\times n\times D}$.  
\item \textbf{Aggregation Stage:} Aggregate $\hat{F}$ with SeqGeM and SeqVLAD into $V \in \mathbb{R}^{C\times D}$.  

\end{itemize}

\subsection*{B. Encoder Stage:} 

We adopt CCT384 as our encoder backbone to extract initial features. Given a sequence $S=\{s_i\}_{i=1}^L$ with $s_i \in \mathbb{R}^{H\times W\times 3}$, the encoder maps each frame into a tokenized feature representation.
\[
    F = \{f_t\}_{t=1}^{L} \in \mathbb{R}^{L\times n\times D},
\]

\noindent The initial encoder output is arranged as $n\times D$, we reinterpret it as $D\times H\times W$. We then flatten $H\times W$ and permute the tensor back to $n \times D$ but with a reordered representation.

\subsection*{C. Recurrent-DTE Stage:} 

\noindent\textbf{DTE:} We choose DTE \cite{Zhu_2021_deformable} for its deformable attention computation, where each query's attention is restricted to its reference point's neighboring keypoints. Following \cite{Zhu_2021_deformable, Li_2022_bevformer}, we overlay each feature map with a uniform 2D reference point grid and learn per query offsets to dynamically select sampling locations. In our setting, we use a single feature level instead of the multi-scale design in the original DTE \cite{Zhu_2021_deformable} to reduce compute cost and keep our model architecture simple. Furthermore, unlike works that employ DTE for 3D perception \cite{Li_2022_bevformer, Xu_2023_bev360}, we restrict our design to 2D feature modeling. This avoids the additional computation needed for 3D spatial modeling and needing known camera intrinsics, making our method deployable across diverse hardware setups.

\noindent\textbf{Recurrent-DTE:} To capture temporal dependencies across frames of $S$, we introduce a recurrent mechanism that, at frame index $t$, uses the previous frame’s DTE output $\hat{f}_{t-1}$ together with the current frame’s encoder features $f_t$. Here, $\hat{f}_{t-1}$ acts as a hidden state shaped by iterations $\{1,\ldots,t-1\}$. At iteration $t$, we set $\hat{f}_{t-1}$ as $Q_t$ and $f_t$ as $K_t$ and  $V_t$. Because each iteration of Recurrent-DTE relies only on the output of the previous one, the module is seq-length agnostic. For initialization, $Q_{t=1}$ is obtained by adding a learnable positional embedding $\Delta \in \mathbb{R}^{n \times D}$ to $f_{t=1}$,  $\Delta$ is applied only at $t=1$.

\begin{equation}
    \begin{gathered}
        \hat{f}_t=\mathrm{DTE}(Q_t=\hat{f}_{t-1},\,K_t=V_t=f_t) \\
        Q_{t=1}=f_{t=1}+\Delta
    \end{gathered}
\end{equation}

\noindent This design differs fundamentally from STformer \cite{Zhao_2024_attention}, where spatial and temporal information are learned by separate transformer encoders. A simple equivalent formulation of STformer using our notation is 

\begin{equation}
    \{\hat{f}_t\}_{t=1}^{L}
    =
    \mathcal{T}_{\mathrm{temporal}}
    \Big(
        \mathcal{T}_{\mathrm{spatio}}(\{f_t\}_{t=1}^{L})
    \Big),
\end{equation}

\noindent where $\mathcal{T}_{(\cdot)}$ denotes a transformer encoder, with its modeling scope indicated by the subscript. In our approach, temporal information is instead implicitly captured within the same transformer operation as spatial modeling. Recurrent-DTE therefore unifies spatio-temporal modeling within a single module.

\begin{equation}
    \hat{f}_t
    =
    \mathcal{T}_{\mathrm{spatio-temporal}}(\hat{f}_{t-1},f_t)
\end{equation}

\subsection*{D. Aggregation Stage}
We concatenate per-iteration outputs \( \{\hat{f}_{t=1}, \dotsc, \hat{f}_{t=L}\} \) into the tensor \( \hat{F} \in \mathbb{R}^{L\times n\times D} \) and permute \(\hat{F}\) to become \(\hat{F}' \in \mathbb{R}^{n\times L\times D}\) (treating tokens as the batch dimension) and apply SeqGeM \cite{Berton_2024_jist}, a learnable mean pooling over the temporal axis, collapsing \(L\) to 1: 
\begin{equation}
\tilde F = \mathrm{SeqGeM}(\hat{F}') \in \mathbb{R}^{1\times n\times D}
\end{equation}
We then feed \(\tilde F\) into SeqVLAD \cite{Mereu_2022_seqvlad}, which aggregates the \(n\) \(D\)-dimensional embeddings into a descriptor
\(V \in \mathbb{R}^{C\times D}\). Both SeqGeM and SeqVLAD are flexible to the value of \(L\).

\section*{Experiment Setup \& Implementation Details}
\begin{table}[ht]
\captionsetup{font=sc}
\captionsetup{font={scriptsize, sc, stretch=1.3}, justification=centering}
\caption{Dataset details specifying the number of images in the database and query sets.}
\label{tab:datasets}
\fontsize{8pt}{8pt}\selectfont
\setlength{\tabcolsep}{10pt} % comfortable column spacing
\centering
\begin{tabular}{@{}lcc@{}}
\toprule
\textbf{Dataset} & \textbf{Split} & \textbf{Database / Queries} \\
\midrule
\multirow{2}{*}{NordLand} 
  & Train set & 15000 / 15000 \\
  & Test set  & 3000 / 3000 \\
\midrule
\multirow{2}{*}{Oxford-Easy} 
  & Train set & 3619 / 3962 \\
  & Test set  & 3632 / 3921 \\
\midrule
\multirow{2}{*}{Oxford-Hard} 
  & Train set & 2322 / 2585 \\
  & Test set  & 2970 / 2920 \\
\midrule
\multirow{1}{*}{NuScenes} 
  & Test set  & 4500 / 4000 \\
\bottomrule
\end{tabular}
\end{table}

\subsection*{A. Dataset}

\noindent\textbf{NordLand:} The Nordland dataset \cite{Sunderhauf_2013_nordland} comprises a collection of images captured during rail journeys across four seasons. We use the Summer-Winter pair for training, and the Spring-Fall pair for testing. 

\noindent\textbf{Oxford RobotCar:} The Oxford RobotCar dataset \cite{Maddern_2017_RobotCar,Maddern_2020_RobotCarRTK} provides large-scale urban driving sequences under diverse conditions. We construct two sub-datasets: \emph{Oxford-Easy}, following the Oxford2 split in \cite{Zhao_2024_attention}, which uses all six route sections with 2 meter frame separation; and \emph{Oxford-Hard}, built from day–night traversals with disjoint route segments (sections 1–2 for training, 3–4 for testing) and 1 meter frame separation. Oxford-Hard is more challenging due to both route mismatch and severe lighting changes between training and test sets.

\noindent\textbf{NuScenes:} 
The NuScenes dataset \cite{Caesar_2020_NuScenes} is a large-scale autonomous driving benchmark collected in urban environments. We select the Singapore scenes to capture urban day–night traversals under extreme lighting transitions. We split scenes into day and night and treat pairs with less than 30° of angular difference as positives, following \cite{Ge_2024_bev2pr}. To assess \textit{cross-dataset generalization}, we evaluate a model trained on Oxford-Easy directly on the NuScenes dataset.  

\subsection*{B. Implementation Detail}

\noindent\textbf{Model hyperparameters:} For the DTE, we use 8 attention heads, 8 sampling points, 2 levels (with only 1 feature level input), a dropout rate of 0.1, and 64 VLAD clusters.

\noindent\textbf{Training:} Following SeqVLAD \cite{Mereu_2022_seqvlad}, we initialize our CCT encoder from its authors' pretrained weights \cite{Hassani_2021_compact} and use the ADAM optimizer \cite{Kingma_2014_adam} with a learning rate of $1\times10^{-5}$. Similar to \cite{Garg_2021_seqnet, Mereu_2022_seqvlad}, we adopt the triplet loss from NetVLAD \cite{Arandjelovic_2016_netvlad}, selecting one non-trivial positive sample from descriptor space and five hardest negative samples from GNSS ground truths. During training, we set input image resolution to $384\times384$, batch size to four, sequence length to five, and triplet margin to 0.1. Training stops early if Recall@5 does not improve for 5 straight epochs. 

\subsection*{C. Baseline Seq-VPR Methods}
As the choice of encoder backbone greatly influences performance and efficiency, we first compare Adapt-STformer (ours) with four representative Seq-VPR baselines without a foundational encoder backbone. Namely, we experiment with SeqNet \cite{Garg_2021_seqnet}, SeqVLAD \cite{Mereu_2022_seqvlad}, STformer \cite{Zhao_2024_attention}, and JIST \cite{Berton_2024_jist}. In section \ref{SOTA_investigation}, we evaluate Adapt-STformer against SOTA VPR methods \cite{lu2024cricavpr, 10884025} that utilize a foundational encoder backbone. For a fairer comparison with such methods trained on much larger datasets, we also analyze their performance under inference time constraints, which produces further insights on the value of Adapt-STformer's efficiency.

\noindent\textbf{Evaluation Criteria.}  
Following prior work, we measure VPR performance by Recall@K, which counts a query correct if at least one of the top-K retrieved candidates lies within 10 m for Oxford RobotCar and NuScenes, or within $\pm$1 frame for Nordland.  

% \textbf{Comparison with ViT-based VPR:}  
% To contextualize our contributions, we additionally compare against recent transformer-based methods that employ ViT foundation backbones \cite{lu2024cricavpr, 10884025}. Unlike Seq-VPR baselines and our method, which use lighter backbones, these approaches achieve strong performane largely by leveraging heavy ViT encoders. While such models are substantially more compute- and memory-intensive, our results show that Adapt-STformer attains competitive performance with far lower resource demands. This separation highlights the benefits of our spatio-temporal design independent of backbone scale, while also enabling a fair assessment against state-of-the-art ViT-based methods.

% ---------- Preamble additions ----------
\newcommand{\best}[1]{\textbf{{#1}}}
\newcommand{\second}[1]{\underline{#1}} 
% ----------------------------------------

\begin{table*}[t]
\centering
\label{tab:results}
\captionsetup{font=sc}
\captionsetup{font={scriptsize, sc, stretch=1.3}, justification=centering, labelsep=newline}

\caption{Adapt-STformer is compared with four Seq-VPR baselines in terms of recall, memory, and inference time per sequence. Best results are shown in \textbf{bold}, second-best in \underline{underline}. We don't quantify PCA impact on resource usage because it is post descriptor extraction. Public checkpoint(pc)}

\label{tab:results}
\scriptsize
\begin{adjustwidth}{0cm}{0cm}
\setlength{\tabcolsep}{2.1pt}
\renewcommand{\arraystretch}{1.25}
\begin{tabular}{@{}clc*{4}{ccc}cc@{}}
\toprule
\multirow{2}{*}{\textbf{\makecell{Descriptor\\Dimension}}} & \multirow{2}{*}{\textbf{ Method}} & \multirow{2}{*}{\textbf{Backbone}} &
\multicolumn{3}{c}{\textbf{NuScenes}} &
\multicolumn{3}{c}{\textbf{Oxford-Hard}} &
\multicolumn{3}{c}{\textbf{Oxford-Easy}} &
\multicolumn{3}{c}{\textbf{Nordland}} &
\multicolumn{2}{c}{\textbf{Resource Usage}} \\
\cmidrule(lr){4-6}\cmidrule(lr){7-9}\cmidrule(lr){10-12}\cmidrule(lr){13-15}\cmidrule(l){16-17}
& & &
R@1 $\uparrow$ & R@5 $\uparrow$ & R@10 $\uparrow$ &
R@1 $\uparrow$ & R@5 $\uparrow$ & R@10 $\uparrow$ &
R@1 $\uparrow$ & R@5 $\uparrow$ & R@10 $\uparrow$ &
R@1 $\uparrow$ & R@5 $\uparrow$ & R@10 $\uparrow$ &
Mem (MB) $\downarrow$ & Time (s) $\downarrow$ \\
\midrule
\multirow{3}{*}{\textbf{24576}}
  & Seqvlad          & CCT384 & 0.3236 & 0.4327 & 0.5142 & 0.4709 & 0.5601 & 0.6151 & 0.8275 & 0.9249 & 0.9541 & 0.9603 & \best{0.9947} & \second{0.9947} & \best{177.66} & \second{0.0181} \\
  & STformer $^{\text{pc}}$         & CCT384 & \second{0.4461} & \second{0.5319} & \second{0.5945} & \second{0.5754*} & \second{0.6548*} & \second{0.7011*} & \second{0.8488} & \second{0.9323} & \second{0.9633} & \second{0.9697} & \best{0.9947} & \best{0.9950} & 276.55 & 0.0277 \\
  & Ours       & CCT384 & \best{0.5496} & \best{0.7035} & \best{0.7622} & \best{0.6940} & \best{0.7706} & \best{0.7969} & \best{0.8854} & \best{0.9528} & \best{0.9700} & \best{0.9763} & \second{0.9943} & \second{0.9947} & \second{180.39} & \best{0.0178} \\
\cmidrule(lr){1-17}
\multirow{4}{*}{\makecell{\textbf{4096}\\(PCA)}}
  & Seqnet           & NetVLAD & 0.0827 & 0.1575 & 0.2031 & 0.3064 & 0.4505 & 0.5228 & 0.5363 & 0.7467 & 0.8241 & 0.7943 & 0.9013 & 0.9317 & 676.09 & \best{0.0014} \\
  & Seqvlad    & CCT384  & 0.3236 & 0.4327 & 0.5146 & 0.4603 & 0.5550 & 0.6151 & 0.8280 & 0.9251 & 0.9557 & 0.9567 & \best{0.9947} & \second{0.9947} & \best{177.66} & 0.0181 \\
  & STformer $^{\text{pc}}$   & CCT384  & 0.4461 & 0.5319 & 0.5949 & \second{0.5648*} & \second{0.6453*} & \second{0.6968*} & \second{0.8534} & \second{0.9372} & \second{0.9667} & \best{0.9700} & \best{0.9947} & \best{0.9950} & 276.55 & 0.0277 \\
  & Ours & CCT384  & \best{0.5488} & \best{0.7031} & \best{0.7622} & \best{0.6874} & \best{0.7659} & \best{0.7918} & \best{0.8882} & \best{0.9544} & \best{0.9710} & \second{0.9610} & \second{0.9923} & 0.9937 & \second{180.39} & \second{0.0178} \\
\cmidrule(lr){1-17}
\multirow{4}{*}{\makecell{\textbf{512}\\(PCA)}}
  & JIST $^{\text{pc}}$ & ResNet  & 0.2047 & 0.3146 & 0.3713 & 0.4423 & 0.5405 & 0.6139 & 0.5868 & 0.7416 & 0.8134 & 0.8210 & 0.9200 & 0.9450 & \best{159.65} & \best{0.0053} \\
  & Seqvlad    & CCT384  & 0.3146 & 0.4236 & 0.5024 & 0.4383 & 0.5330 & 0.6049 & 0.8236 & 0.9244 & 0.9564 & \second{0.9567} & \best{0.9947} & \second{0.9947} & \second{177.66} & 0.0181 \\
  & STformer $^{\text{pc}}$   & CCT384  & \second{0.4445} & \second{0.5370} & \second{0.5969} & \second{0.5377*} & \second{0.6198*} & \second{0.6720*} & \second{0.8513} & \second{0.9369} & \second{0.9667} & \best{0.9637} & \second{0.9933} & \best{0.9950} & 276.55 & 0.0277 \\
  & Ours & CCT384  & \best{0.5551} & \best{0.7091} & \best{0.7717} & \best{0.6516} & \best{0.7459} & \best{0.7765} & \best{0.8870} & \best{0.9554} & \best{0.9731} & 0.9490 & 0.9900 & 0.9943 & 180.39 & \second{0.0178} \\
\bottomrule
\end{tabular}
\end{adjustwidth}
\vspace{0.3em}
\scriptsize STformer is evaluated using its pc since the full training code is unavailable. For most datasets, the STformer's pc was trained on the same training sets as ours. However, for Oxford-Hard ($^*$), the training data differs, so we finetune STformer to ensure a fair comparison; these results are reported but not included in our analysis. JIST is also evaluated using a pretrained checkpoint (pc) because its training procedure differs from ours.
\end{table*}

\section*{Results}

\subsection*{A. Recall comparison}  

We report the evaluation results of Adapt-STformer and Seq-VPR baselines in Table~\ref{tab:results}. We calculate Recall@K for top K values of 1, 5, and 10 and measure the memory usage and duration of one forward pass for each model.

\noindent\textbf{Nordland and Oxford-Easy:}
SeqVLAD, STformer, and our method achieve near-perfect recall on Nordland. This can be attributed to Nordland’s minor lighting variations between the query and the database, making it an easy dataset. On the Oxford-Easy dataset, we achieve a +3\% R@1 gain over STformer. While Oxford-Easy presents more lighting challenges than Nordland, its train and test sets share overlapping locations, simplifying the task. 

\noindent\textbf{Oxford-Hard and NuScenes:} Substantial gains are observed on harder benchmarks: Oxford-Hard (poor lighting, rain, and unseen sections) and NuScenes (poor lighting and cross-domain train-test splits). On Oxford-Hard, we achieve approximately +20\% gains in R@1,5,10 over SeqVLAD. For NuScenes, we observe +10\%,17\%,17\% improvements in R@1,5,10, respectively, over STformer. 

\noindent\textbf{PCA:} We standardise descriptor sizes via PCA \cite{Jegou_2012_pca} for fairness. The relative performance ranking of the methods remains unchanged.

\subsection*{B. Resource Usage}  

Computational efficiency is critical for practical deployment of VPR systems and a central focus of our work. Table~\ref{tab:results} reports the overall resource usage of all methods. In table~\ref{tab:time_breakdown_reorg}, we make a per-module inference times comparison between Adapt-STformer (ours) and STformer, the most architecturally similar baseline. All experiments were conducted using PyTorch~\cite{Paszke_2019_pytorch} on an NVIDIA Tesla V100.

\noindent\textbf{Transformer Modules:} For spatially attending to a single frame, Adapt-STformer's DTE module is 71.3\% faster than the non-deformable transformer encoder in STformer. Also when spatio-temporally processing an entire query, our Recurrent-DTE architecture uses 88.1\% less time than STformer's additional temporal transformer encoder.

\noindent\textbf{Aggregation Modules:} While both methods employ SeqVLAD for sequence aggregation, Adapt-STformer first applies SeqGeM to collapse the aggregation input shape from ${L \times n \times D}$ into $1 \times n \times D$,  introducing negligible overhead while accelerating the SeqVLAD operation by 56.8\%.

\noindent\textbf{Overall Comparison:} Overall, Adapt-STformer extracts descriptors 36\% faster per sequence and requires 35\% less memory than STformer (Table~\ref{tab:time_breakdown_reorg}). Notably, it has faster extraction than SeqVLAD, \emph{a non-transformer method.}

\begin{table}[t]
\centering
\captionsetup{font=sc}
\captionsetup{font={scriptsize, sc, stretch=1.3}, justification=centering}
\caption{Per-module inference time breakdown and overall resource usage comparison between our method and STformer.}
\label{tab:time_breakdown_reorg}
\scriptsize
\setlength{\tabcolsep}{5pt}
\begin{tabular}{@{}lccccc@{}}
\toprule
\multirow{2}{*}{Method} & \multicolumn{4}{c}{Inference Time (ms)} & \multirow{2}{*}{\makecell{Memory$\downarrow$\\(MB)}} \\
\cmidrule{2-5}
& \textbf{Spatial Attn.} $\downarrow$ & \textbf{Spatio-Temp}$\downarrow$ & \textbf{Aggr.}$\downarrow$ & \textbf{Overall}$\downarrow$ & \\
\midrule
STformer & 1.71 & 23.02 & 3.98 & 27.7 & 276.55 \\
Ours     & \best{0.49} & \best{2.73} & \best{1.72} & \best{17.8} & \best{180.39} \\
\midrule
\% Improve & \best{71.3\%} & \best{88.1\%} & \best{56.8\%} & \best{35.7\%} & \best{34.8\%} \\
\bottomrule
\end{tabular}

\vspace{3pt} % adjust spacing as needed
\scriptsize \textbf{Note:} Our encoder backbone takes approx. 14 ms, accounting for the gap between module times and total inference time.
\end{table}

\begin{table}[t]
\centering
\captionsetup{font=sc}
\captionsetup{font={scriptsize, sc, stretch=1.3}, justification=centering}
\caption{Ablation study of seq-lengths \(L\) and spatio-temporal methods. We compare: (i) \textbf{Recurrent-TE} (non-deformable Transformer Encoder with recurrence), (ii) \textbf{DTE+TT} (DTE with a separate temporal transformer), (iii) \textbf{DTE Only} (ours withoutrecurrence), and (iv) \textbf{Recurrent-DTE} (ours).}

\label{tab:ablation_combined}
\scriptsize
\setlength{\tabcolsep}{2.5pt}
\begin{tabular}{@{}c l c c c c@{}} % <-- first column centered
\toprule
\multirow{2}{*}{\textbf{\(L\)}} &
\multirow{2}{*}{\textbf{Spatio-Temp. Method}} &
\multicolumn{1}{c}{\textbf{Oxford-Hard}} &
\multicolumn{1}{c}{\textbf{NuScenes}} &
\multicolumn{2}{c}{\textbf{Resource Usage}} \\
\cmidrule(lr){3-3}\cmidrule(lr){4-4}\cmidrule(l){5-6}
& & R@5 $\uparrow$ & R@5 $\uparrow$ & Mem(MB) $\downarrow$ & Time(s) $\downarrow$ \\
\midrule

\multirow{4}{*}{\textbf{1}}
  & Recurrent-TE & 0.4776 & 0.1173 & \second{98.15} & 0.0120 \\
  & DTE + TT & \best{0.7368} & 0.5106 & 106.02 & 0.0122 \\
  & DTE Only (non-recurrent) & 0.7306 & \second{0.5283} & \best{88.71} & \best{0.0111} \\
  & Recurrent-DTE (Ours) & \second{0.7341} & \best{0.6224} & \best{88.71} & \second{0.0115} \\
\midrule

\multirow{4}{*}{\textbf{3}}
  & Recurrent-TE & 0.4552 & 0.1303 & \second{136.75} & 0.0135 \\
  & DTE + TT & \second{0.7384} & 0.4764 & 152.03 & 0.0177 \\
  & DTE Only (non-recurrent) & 0.7255 & \second{0.4898} & \best{134.39} & \second{0.0126} \\
  & Recurrent-DTE (Ours) & \best{0.7549} & \best{0.6681} & \best{134.39} & \best{0.0128} \\
\midrule

\multirow{4}{*}{\textbf{5}}
  & Recurrent-TE & 0.4085 & 0.111 & \second{182.75}  & 0.0181 \\
  & DTE + TT & \second{0.7341} & 0.4677 & 198.16 & 0.0267 \\
  & DTE Only (non-recurrent) & 0.7247 & \second{0.4693} & \best{180.39} & \best{0.0176} \\
  & Recurrent-DTE (Ours) & \best{0.7706} & \best{0.7039} & \best{180.39} & \second{0.0178} \\
\bottomrule
\end{tabular}

\vspace{3pt} % adjust spacing as needed
\scriptsize \textbf{Disclaimer:} Recurrent-TE and the DTE + TT ablation baselines were integrated with minimal tuning. 

\end{table}

\subsection*{D. Ablation Study}  

We conduct ablation studies on Oxford-Hard and NuScenes to assess the effects of sequence length $L$ and the proposed Recurrent-DTE. All models were trained on $L=5$ and evaluated at $L=3$ and $L=1$, demonstrating flexibilty to shorter seq-lengths under limited inference resources \cite{Malone_2024_modulating}

\noindent\textbf{Sequence Lengths:}  
As shown in Table~\ref{tab:ablation_combined}, performance increases with longer sequences. On Oxford-Hard, our Recurrent-DTE rises from 0.7341 ($L=1$) to 0.7706 ($L=5$), while on NuScenes it improves from 0.6224 to 0.7039. This demonstrates the ability of our approach to exploit additional temporal context as $L$ grows effectively.  

\noindent\textbf{Spatio-Temporal Methods:}  
Table~\ref{tab:ablation_combined} further compares different spatio-temporal methods. In the non-recurrent variant (DTE Only), each frame is processed independently, with $Q_t,K_t,V_t=f_t$, yielding no cross-frame interaction. Our Recurrent-DTE consistently outperforms this baseline across all sequence lengths, with gains of up to 18\% (NuScenes, $L=3$), while requiring no additional memory and only negligible runtime overhead. We also evaluate using a separate temporal transformer layer (DTE + TT), similar to \cite{Zhao_2024_attention,10884025}. Although competitive on Oxford-Hard ($L=1$: 0.7368 vs. 0.7341), this approach is substantially weaker on NuScenes ($L=3$: 0.4764 vs. 0.6681), while being 32\% slower (0.0177s vs. 0.0128s) and 12\% more memory usage (152.03MB vs. 134.39MB). These results demonstrate that our recurrent-DTE is efficient and effective at spatio-temporal modeling.  

\noindent\textbf{DTE Module:}  
Finally, replacing our deformable transformer encoder with a non-deformable variant (Recurrent-TE baseline) degrades both performance and efficiency. As shown in Table~\ref{tab:ablation_combined}, the deformable design improves recall by +29.4\% on Oxford-Hard ($L=5$: 0.7706 vs. 0.4085) and +492\% on NuScenes ($L=5$: 0.7039 vs. 0.111), while simultaneously lowering memory (180.39MB vs. 182.75MB) and inference time (0.0178s vs. 0.0181s).

\begin{figure}[t]
    \centering
    \begin{subfigure}{\linewidth}
        \centering
        \includegraphics[width=\linewidth]{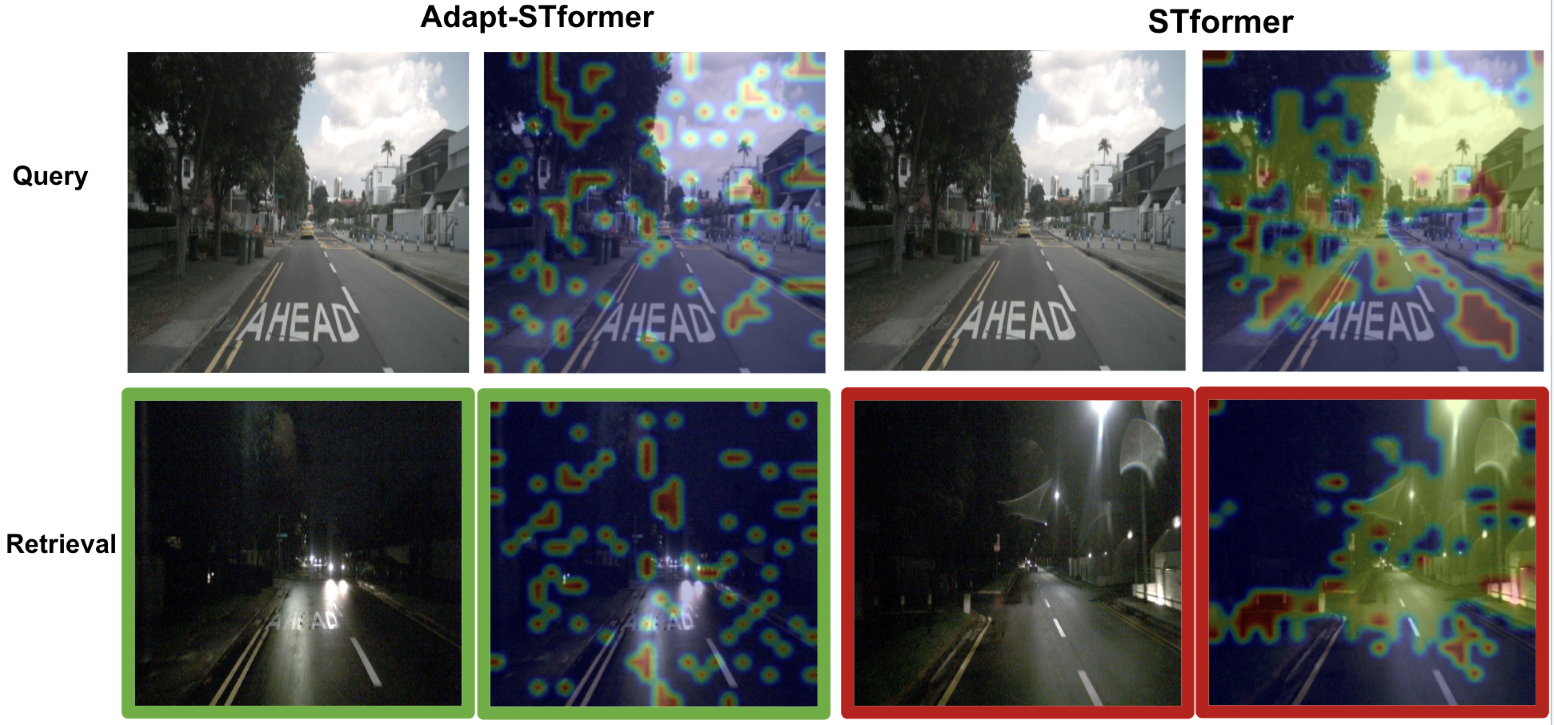}
        \caption{NuScenes Example 1}
        \label{fig:nuscene_ex1}
    \end{subfigure}

    \vspace{0.5em} % optional spacing

    \begin{subfigure}{\linewidth}
        \centering
        \includegraphics[width=\linewidth]{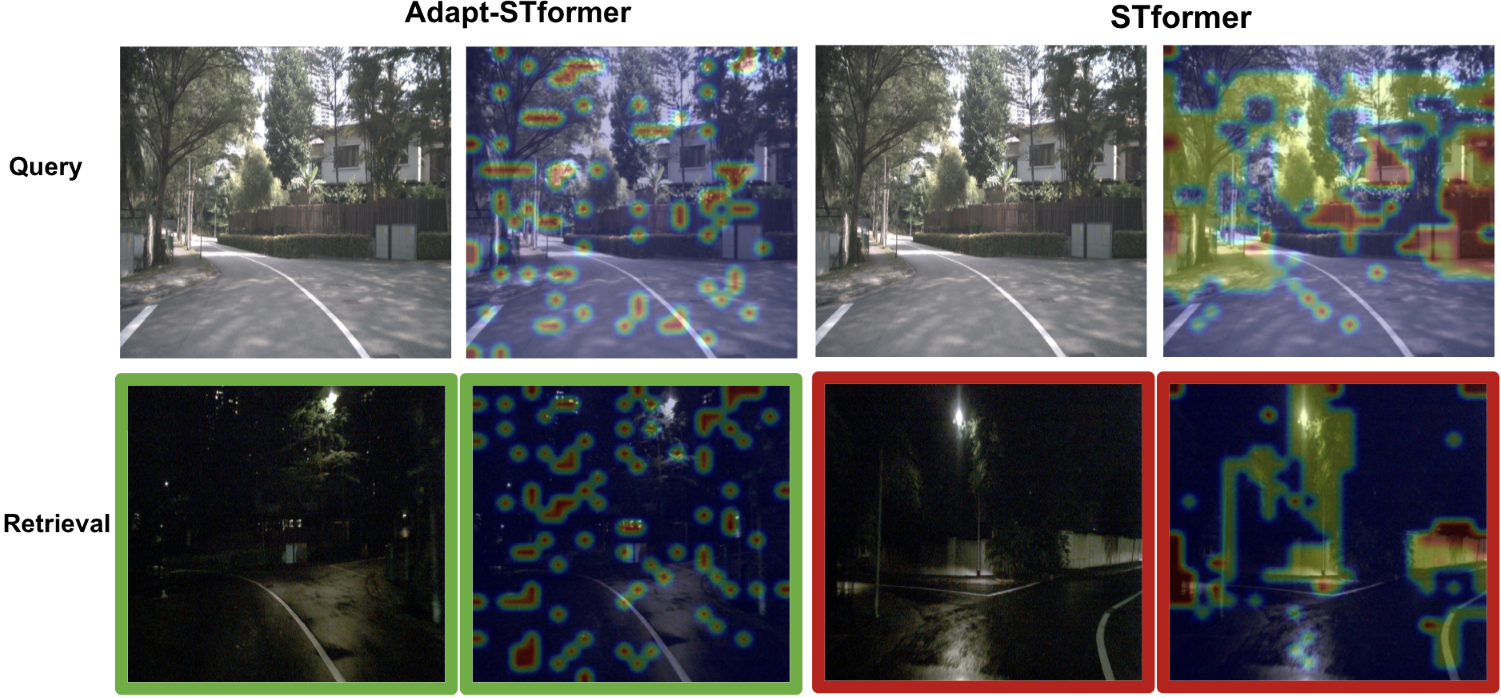}
        \caption{NuScenes Example 2}
        \label{fig:nuscene_ex2}
    \end{subfigure}

\caption{Qualitative comparison of attention maps between our method and STformer on NuScenes. STformer fails in these cases, whereas our method succeeds in VPR matching. In the attention maps, blue denotes low-focus regions and yellow to red gradient denotes high-focus regions by the respective models.}\label{fig:qualitative_results}
\end{figure}

\subsection*{E. Qualititative Results}

Based on the ablation results, DTE is effective at capturing spatial features. Figure~\ref{fig:qualitative_results} shows NuScenes examples where our model retrieves correctly while STformer fails. In both day and night frames, STformer produces smooth, concentrated activations over only a few regions. By contrast, our model displays scattered high-activation patches across the entire frame. This difference, we theorize, arises from the type of attention used. Standard attention, used by STformer, computes attention between every token pair, producing blob-shaped activations around globally salient regions.
In contrast, deformable attention, used by our method, learns a small set of sparse offsets that jump directly to features relative to the query token, yielding scattered activations. Under poor lighting, many features are weak, making it crucial to capture any available geometric cues scattered across the scene. Under poor lighting, it is essential to exploit sparse geometric cues across the scene rather than depend on regions salient only under good illumination.

\begin{figure}[t]
    \centering
    \begin{subfigure}{0.48\linewidth}
        \centering
        \includegraphics[width=\linewidth]{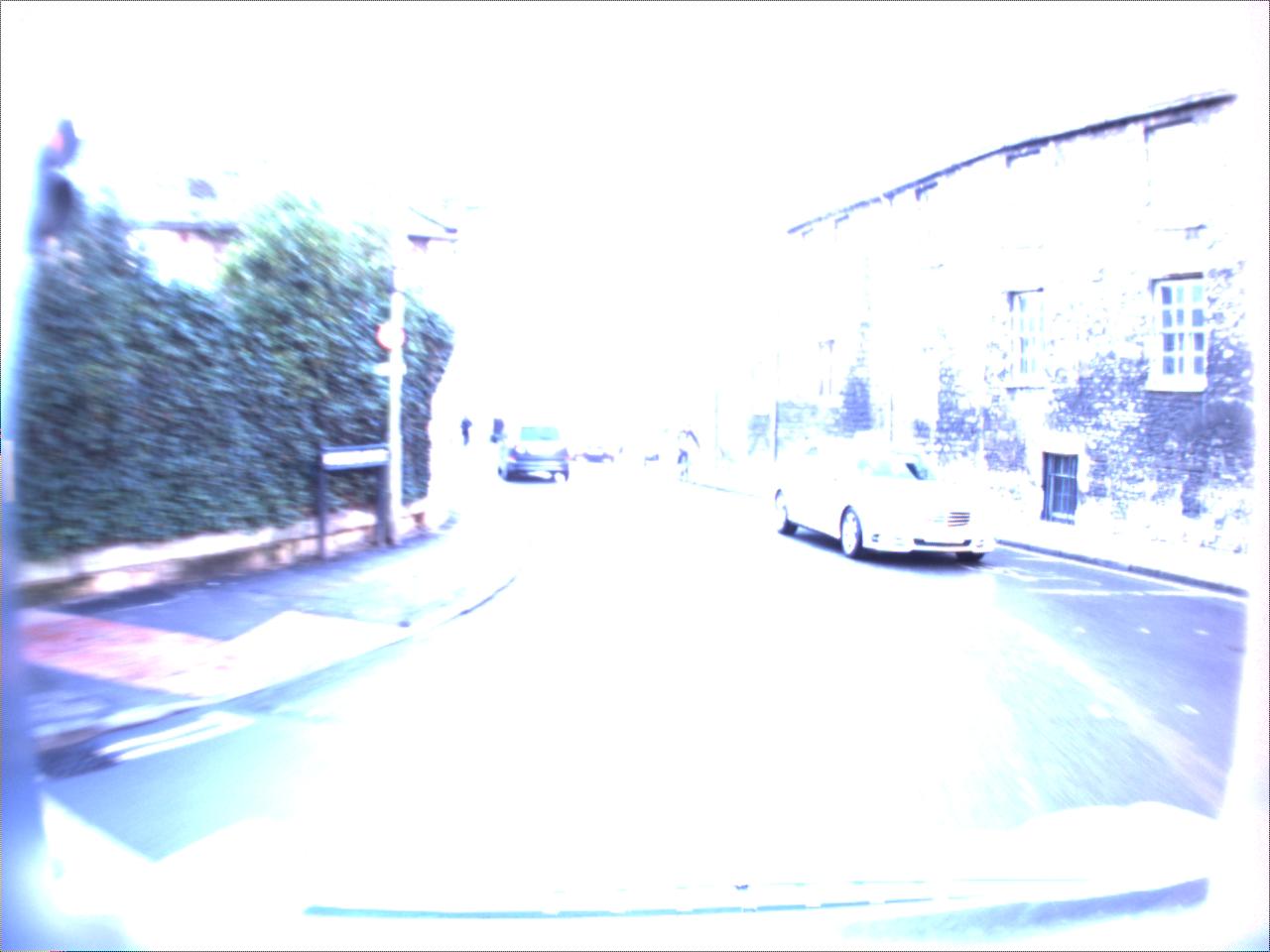}
        \caption{Raw image input}
        \label{fig:refine_a}
    \end{subfigure}
    % \hfill
    \begin{subfigure}{0.48\linewidth}
        \centering
        \includegraphics[width=\linewidth]{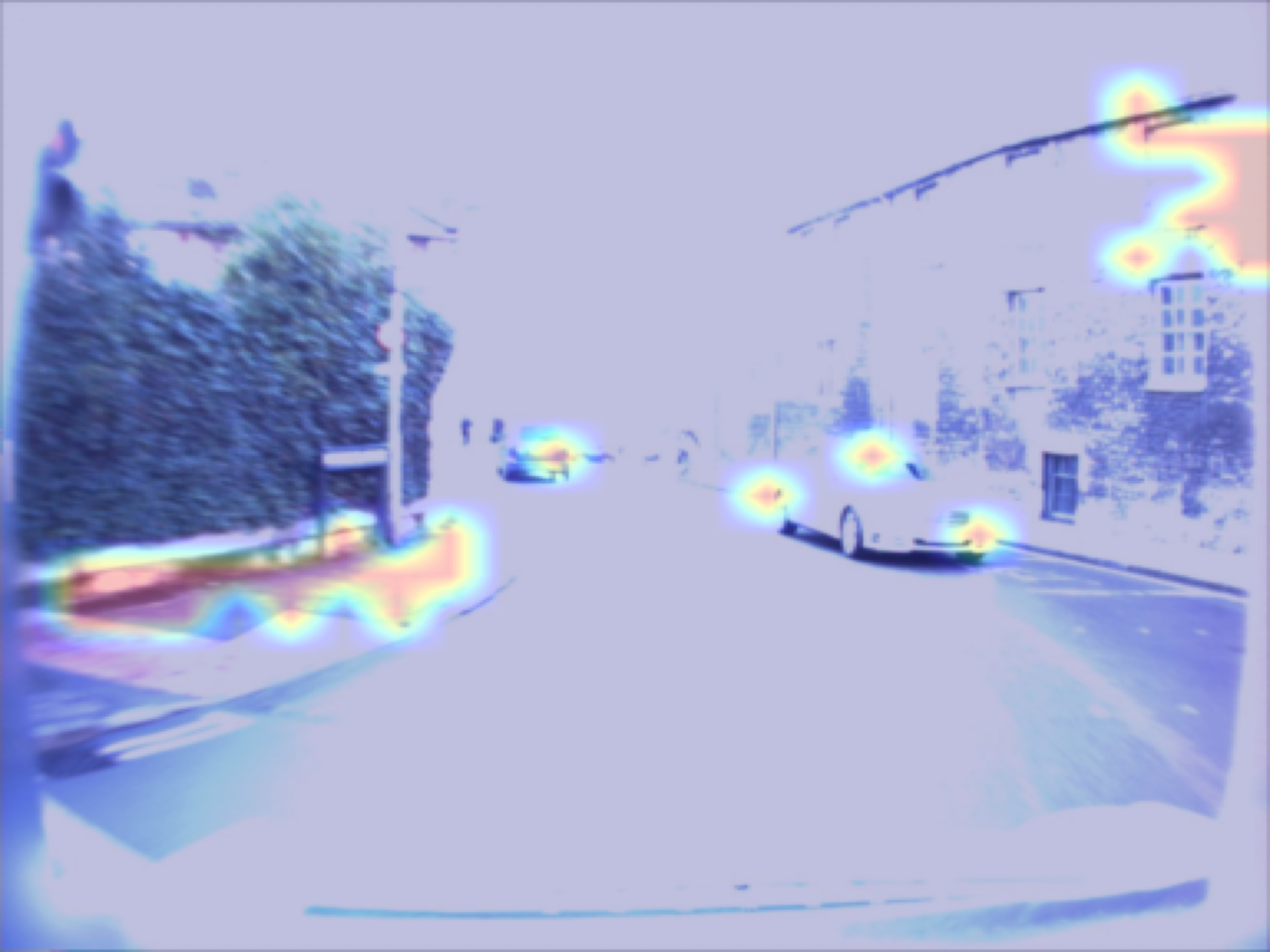}
        \caption{Encoder Selection}
        \label{fig:refine_b}
    \end{subfigure}

    \vspace{0.5em} % small gap between rows

    \begin{subfigure}{0.48\linewidth}
        \centering
        \includegraphics[width=\linewidth]{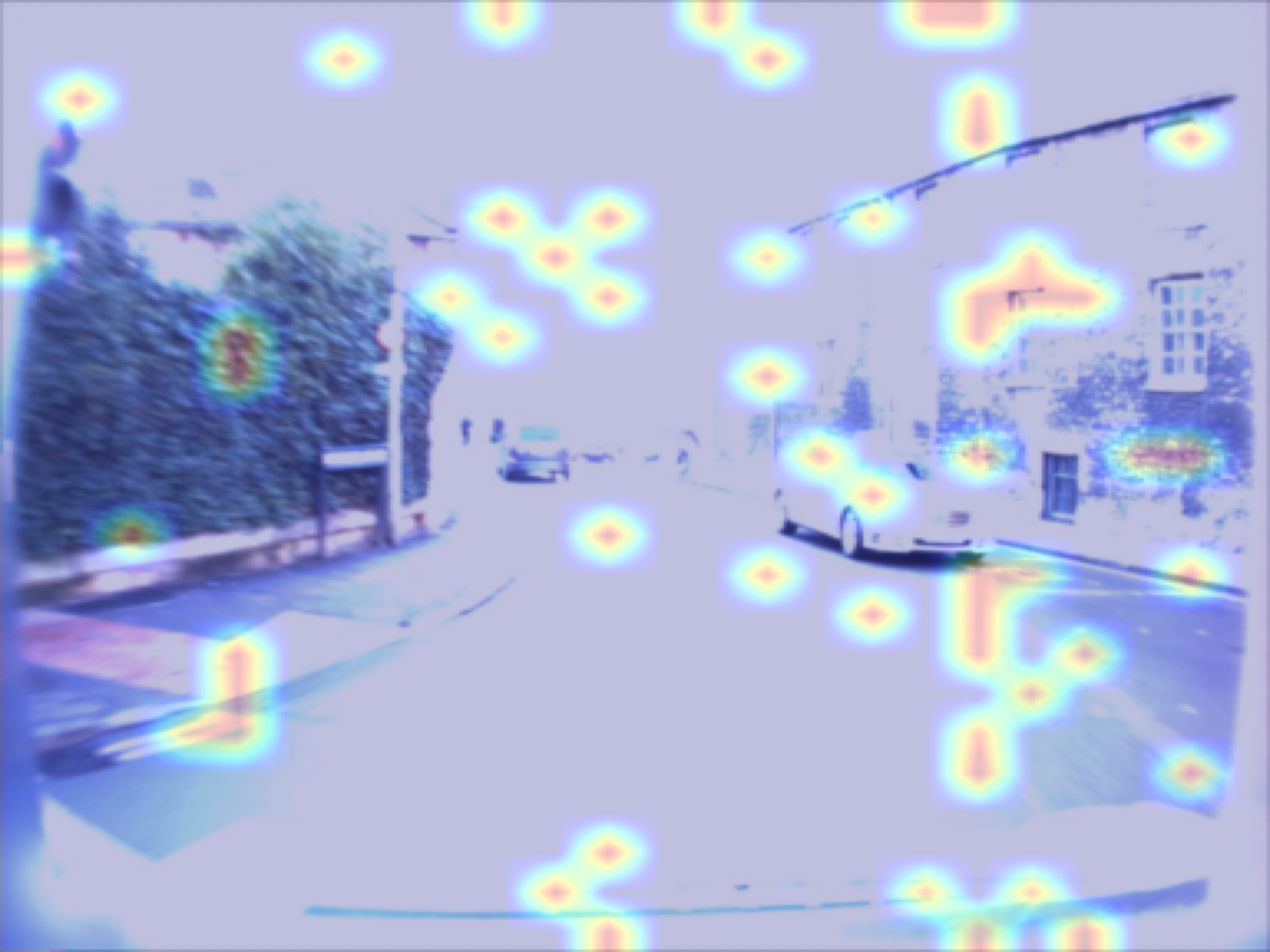}
        \caption{DTE Selection}
        \label{fig:refine_c}
    \end{subfigure}
    % \hfill
    \begin{subfigure}{0.48\linewidth}
        \centering
        \includegraphics[width=\linewidth]{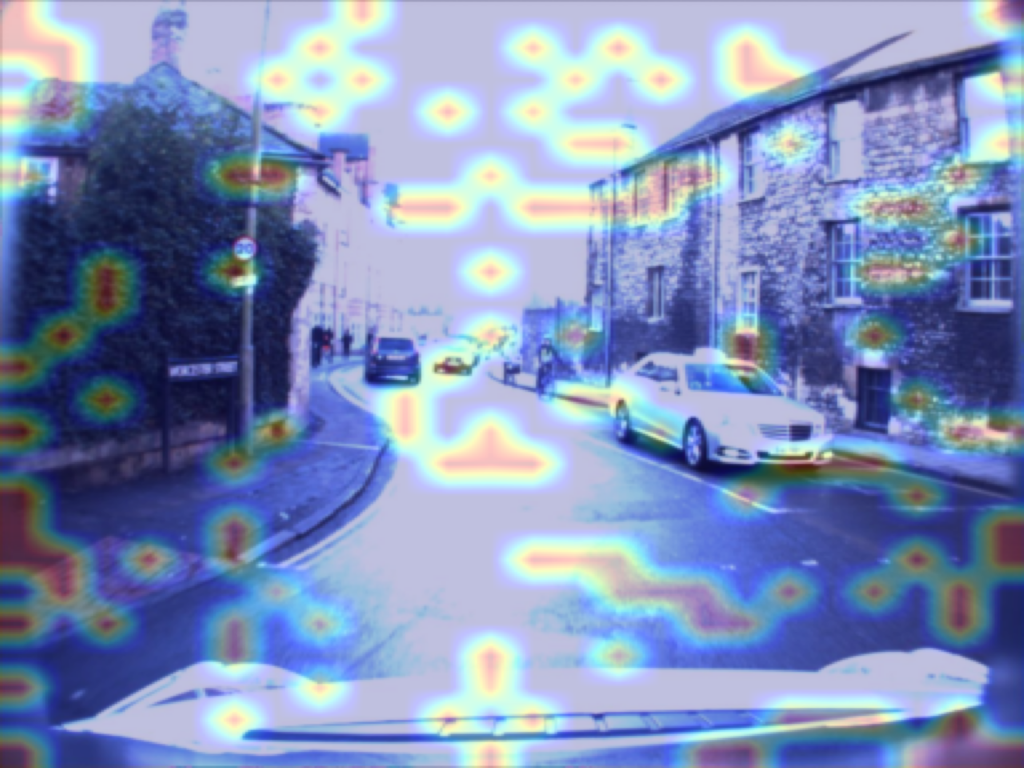}
        \caption{Recurrent-DTE Selection}
        \label{fig:refine_d}
    \end{subfigure}

\caption{Attention Maps Across Adapt-STformer Stages. Subfigs. (b)--(d) show attention activation maps overlaid on the input frame, illustrating model focus at each stage. Color coding follows Fig.~\ref{fig:qualitative_results}.}
\label{fig:refinement}
\end{figure}

\subsection*{F. Attention Maps Across Adapt-STformer Stages}
Figure~\ref{fig:refinement} illustrates the evolution of attention map across the stages of Adapt-STformer when processing a lighting-distorted frame (a) drawn from a sequence. After the encoder stage (b), the model focuses on only a few confident regions while largely ignoring areas affected by lighting distortion.Following DTE processing (c), the model attends to a broader and more discriminative set of VPR-relevant features, emphasizing stable structural cues (e.g., walls, road, vegetation) that remain largely invariant across time and environmental conditions. After the Recurrent-DTE stage (d), cross-frame interactions allow the model to incorporate information from other frames in the sequence. As illustrated in (d), where attention is overlaid on a clearer frame from the same sequence, activation expands to a broader set of stable VPR-relevant cues compared to (c), indicating that the model leverages complementary information across frames to recover features affected by lighting distortions.

\subsection*{G. Comparison with SOTA VPR Methods} \label{SOTA_investigation}
Though Adapt-STformer outperforms Seq-VPR baselines without a foundational encoder backbone, it has yet to be compared against SOTA VPR methods. We benchmark Adapt-STformer against CricaVPR~\cite{lu2024cricavpr} and CaseNet~\cite{10884025}. We also include STformer, our method’s closest architectural counterpart, as best comparable baseline. Besides conventional recall (unlimited time budget), we analyze real-world performance under inference time constraints.

\begin{table*}[!htp]
\captionsetup{font=sc}
\captionsetup{font={scriptsize, sc, stretch=1.3}, justification=centering, labelsep=newline}
\caption{Performance of Adapt-STformer and SOTA VPR Methods. We compute the Recall@5 (R@5), Recall@5 given on-time processing of queries at 36 FPS (R@5-36FPS), the percentage of queries processed on-time (OT\%), and resource usage.}
\label{tab:lat_results}
\scriptsize
\begin{adjustwidth}{0cm}{0cm}
\setlength{\tabcolsep}{4.1pt}
\renewcommand{\arraystretch}{1.25}
\begin{tabular}{@{}l*{4}{cc}cccc@{}}
\toprule
\multirow{2}{*}{\textbf{Method}} &
\multicolumn{2}{c}{\textbf{Nordland}} &
\multicolumn{2}{c}{\textbf{Oxford-Easy}} &
\multicolumn{2}{c}{\textbf{Oxford-Hard}} &
\multicolumn{2}{c}{\textbf{NuScenes}} &
\multirow{2}{*}{OT\% $\uparrow$} &
\multicolumn{3}{c}{\textbf{Resource Usage}} \\
\cmidrule(lr){2-3}\cmidrule(lr){4-5}\cmidrule(lr){6-7}\cmidrule(lr){8-9}\cmidrule(l){11-13}
& R@5 $\uparrow$ & R@5-36FPS $\uparrow$ 
& R@5 $\uparrow$ & R@5-36FPS $\uparrow$
& R@5 $\uparrow$ & R@5-36FPS $\uparrow$
& R@5 $\uparrow$ & R@5-36FPS $\uparrow$
&  & GFLOPs $\downarrow$ & Mem (MB) $\downarrow$ & Time (s) $\downarrow$ \\
\midrule
CricaVPR $^{\text{pc}}$   & 0.9900 & 0.4953 & \underline{0.9803} & 0.4891 & \underline{0.8276} & \underline{0.3822} & \underline{0.8626} & \textbf{0.8622} & 68.36 & 138.62 & 517.77 & 0.029 \\
CaseVPR $^{\text{pc}}$    & 0.9850 & 0.3280 & \textbf{0.9882} & 0.3280 & \textbf{0.8492} & 0.2549 & \textbf{0.8642} & 0.2081 & 33.34 & 503.62 & 641.33 & 0.070 \\
STFormer $^{\text{pc}}$   & \textbf{0.9947} & \underline{0.5153} & 0.9372 & \underline{0.9136} & 0.6453 & 0.3000 & 0.5319 & 0.5154 & \underline{84.78} & \underline{127.74} & \underline{276.55} & \underline{0.028} \\
Ours       & \underline{0.9923} & \textbf{0.9923} & 0.9618 & \textbf{0.9618} & 0.7659 & \textbf{0.7659} & 0.7031 & \underline{0.7031} & \textbf{100.0} & \textbf{102.00} & \textbf{180.39} & \textbf{0.018} \\
\bottomrule
\end{tabular}
\end{adjustwidth}
\vspace{0.3em}
\scriptsize We use pc for CaseVPR and CricaVPR. Both methods were trained on substantially larger datasets than ours; additionally, the CaseVPR pc was further fine-tuned on a dataset equivalent to Oxford-Easy.
\end{table*}

\begin{figure}[t] \centering \includegraphics[width=\columnwidth]{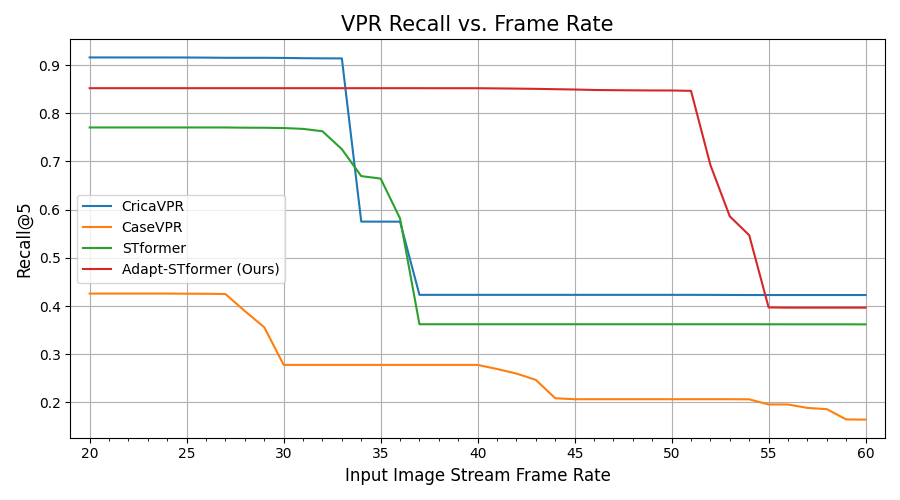} \caption{\textbf{VPR Performance Under Time Constraints.} We analyze the impacts of changing inference time constraints on the performance of Adapt-STformer and other SOTA VPR methods by streaming queries from 20 FPS to 60 FPS.} \label{fig:in_place} \end{figure}

\begin{figure}[t]
    \centering
    \includegraphics[width=1\columnwidth]{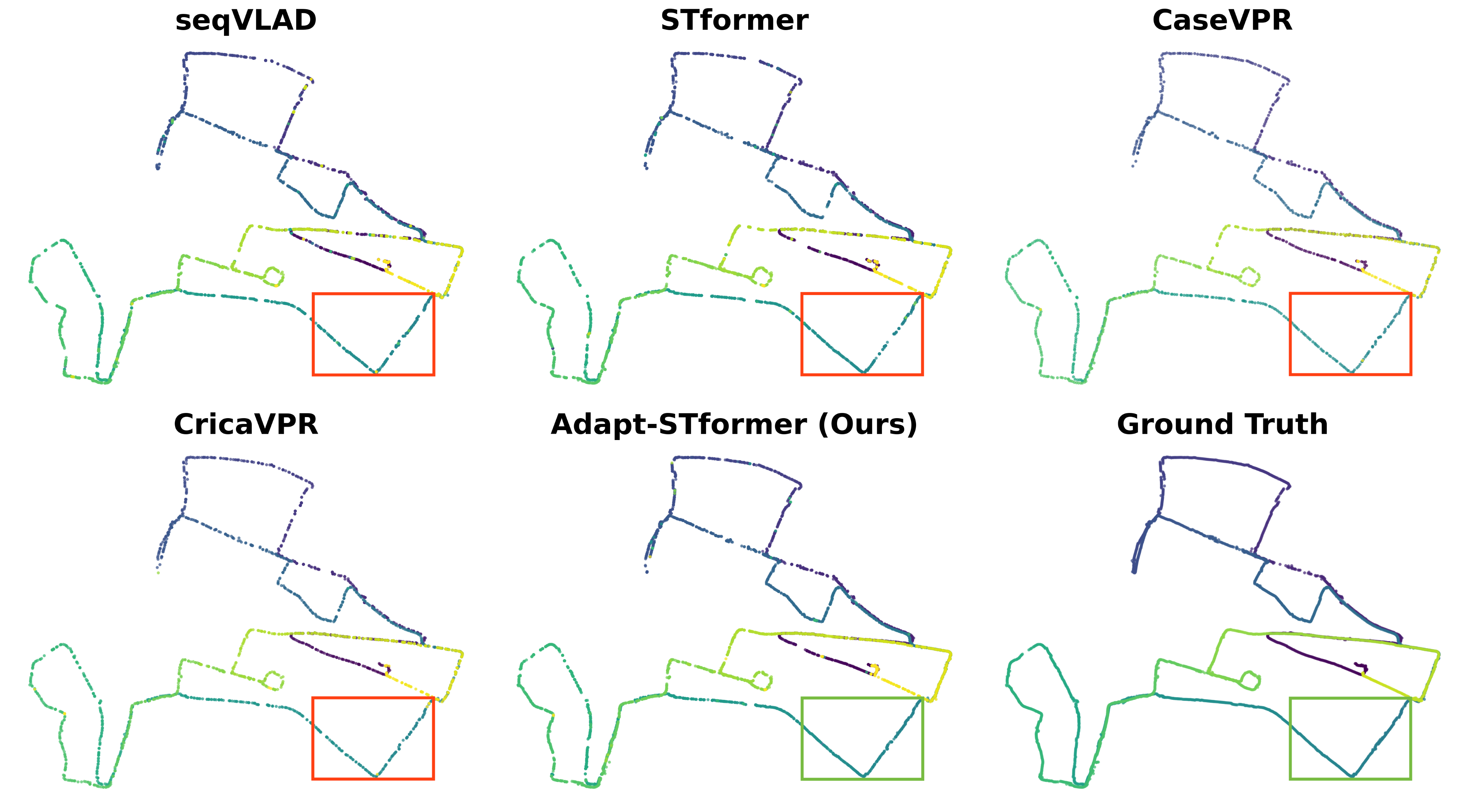}
    \caption{Qualitative trajectories (36 FPS) mappings of various methods on Oxford-Easy,color-coded by frame index. Adapt-STformer aligns closest with GT, due to its high latency-adjusted recall (Table~\ref{tab:lat_results}). We highlight an example region with a green box to show good alignment while red box indicates poor alignment.}
    \label{fig:oxford_easy_mapping}
\end{figure}

We stream queries at 36 FPS—matching the reported Tesla surround-camera rate—for results in Table~\ref{tab:lat_results} and Fig.~\ref{fig:oxford_easy_mapping}. Predictions arriving after the next query are not counted under Recall@K. Table~\ref{tab:lat_results} reports (i) conventional Recall@5, (ii) the percentage of queries processed on time (OT\%), and (iii) Recall@5 over the on-time subset. Fig.~\ref{fig:in_place} further evaluates performance across 20–60 FPS, covering the typical driver-assistance range~\cite{gehrig2024lowlatency}.

\noindent\textbf{Recall under Time Constraints:} 
At 36\,fps, only Adapt-STformer achieves OT\%=100\%, so recall remains equal to its conventional R@5 across all datasets (Table~\ref{tab:lat_results}). 
Against STFormer (the next-fastest), our recall is higher by +47.7\% on Nordland (0.9923 vs.\ 0.5153), 
+4.8\% on Oxford-Easy, 
+46.6\% on Oxford-Hard, 
and +18.8\% on NuScenes (0.7031 vs.\ 0.5154). 
Although DINOv2-based CaseVPR and CricaVPR often lead in conventional R@5, their lower OT\% (33.34\% and 68.36\%) causes large drops in recall under time constraints; 
an exception is NuScenes, where CricaVPR remains best at 0.8622. 
Overall, Adapt-STformer sustains high recall under time pressure while using the fewest GFLOPs, the least memory, and the lowest per-sequence inference latency.

\noindent\textbf{Qualitative Trajectory Mappings:} In Fig.~\ref{fig:oxford_easy_mapping}, we plot predicted query locations on Oxford-Easy. Adapt-STformer produces the smoothest and densest trajectories, closely tracking ground truth. Here, creating visually accurate trajectory mappings requires both high recall and fast inference: methods with low recall (e.g., SeqVLAD) produce fragmented trajectories despite fast speeds, while slower methods (CricaVPR, CaseNet) omit many queries due to latency constraints, likewise degrading trajectory continuity. JIST and SeqNet are omitted as their behavior is similar to SeqVLAD but visually more fragmented.

\noindent\textbf{Practical implication:} When real-time constraints apply, a lighter backbone paired with an efficient spatio-temporal module yields better timely recognition -- higher OT\% and higher recall under time constraints -- while also reducing memory footprint and inference time. This makes our approach a stronger choice for latency-critical VPR applications, even against more powerful alternatives.

\section{Conclusion}

\noindent\textbf{Limitations \& Future Works:} We did not evaluate alternative encoder backbones, such as DINOv2 \cite{Oquab_2023_dinov2}, within Adapt-STformer.
Future work could explore alternative encoder backbones within our framework to preserve its compute efficient design while closing the performance gap with SOTA foundational based VPR models.

\noindent\textbf{Summary:} We present Adapt-STformer, a Seq-VPR framework that delivers competitive performance while remaining compute-efficient and flexible to variable sequence lengths, thanks to its core innovation—the Recurrent-DTE module—which unifies spatio-temporal modeling in a single module. We hope this work provides a foundation for future practical and powerful Seq-VPR research.

\section*{Acknowledgment}
This work was supported by NYU IT High Performance Computing resources, services, and staff expertise. We also thank Somik Dhar for his early contributions. Chen Feng holds concurrent appointments as an NYU Professor and as an Amazon Scholar. This paper describes work performed at NYU and is not associated with Amazon.

\scalefont{0.9}
\bibliographystyle{IEEEtran}
\bibliography{IEEEabrv,refs_abbrev}

\end{document}